\documentclass[lettersize,journal]{IEEEtran}

\usepackage{amsmath,amsfonts}
\usepackage{array}
\usepackage{textcomp}
\usepackage{stfloats}
\usepackage{url}
\usepackage{verbatim}
\usepackage{graphicx}
\usepackage{cite}
\usepackage{algorithm} 
\usepackage{algorithmic}  
\usepackage[algo2e]{algorithm2e}

\usepackage{tabularx}
\usepackage{graphicx}
\usepackage{subfigure}
\usepackage{caption}
\usepackage{amssymb}
\usepackage{array,multirow}
\usepackage{CJKutf8}
\usepackage{booktabs}
\usepackage{caption}
\usepackage{enumitem}
\usepackage{xcolor}
\usepackage{hyperref}       
\usepackage{url}

\long\def\ignore#1{}
\newcommand{\etal}{\emph{et al.}\xspace}

\begin{document}

\title{Improving Compositional Generalization in Math Word Problem Solving}

\author{Yunshi Lan, Lei Wang, Jing Jiang, Ee-Peng Lim
\thanks{Yunshi Lan and Lei Wang contribute equally to this work.}
\thanks{Yunshi Lan is with School of Data Science and Engineering, East China Normal University.
E-mail: yslan@dase.ecnu.edu.cn.}
\thanks{Lei Wang (corresponding author), Jing Jiang and Ee-Peng Lim are with School of Computing and Information System, Singapore Management University.
E-mail: lei.wang.2019@phdcs.smu.edu.sg, jingjiang@smu.edu.sg and eplim@smu.edu.sg.}}

\markboth{Journal of \LaTeX\ Class Files,~Vol.~14, No.~8, August~2021}%
{Shell \MakeLowercase{\textit{et al.}}: A Sample Article Using IEEEtran.cls for IEEE Journals}


\maketitle

\begin{abstract}
Compositional generalization refers to a model's capability to generalize to newly composed input data based on the data components observed during training.
It has triggered a series of compositional generalization analysis on different tasks as generalization is an important aspect of language and problem solving skills. 
However, the similar discussion on math word problems (MWPs) is limited.
In this manuscript, we study compositional generalization in MWP solving. 
Specifically, we first introduce a data splitting method to create compositional splits from existing MWP datasets.
Meanwhile, we synthesize data to isolate the effect of compositions.
To improve the compositional generalization in MWP solving, we propose an iterative data augmentation method that includes diverse compositional variation into training data and could collaborate with MWP methods.
During the evaluation, we examine a set of methods and find all of them encounter severe performance loss on the evaluated datasets.
We also find our data augmentation method could significantly improve the compositional generalization of general MWP methods.
Code is available at \url{https://github.com/demoleiwang/CGMWP}. 
\end{abstract}

\begin{IEEEkeywords}
Math word problem solving, compositional generation, data augmentation.
\end{IEEEkeywords}

\section{Introduction}
\label{sec:intro}

\begin{figure}[!t]
    \centering
    \includegraphics[width=0.47\textwidth]{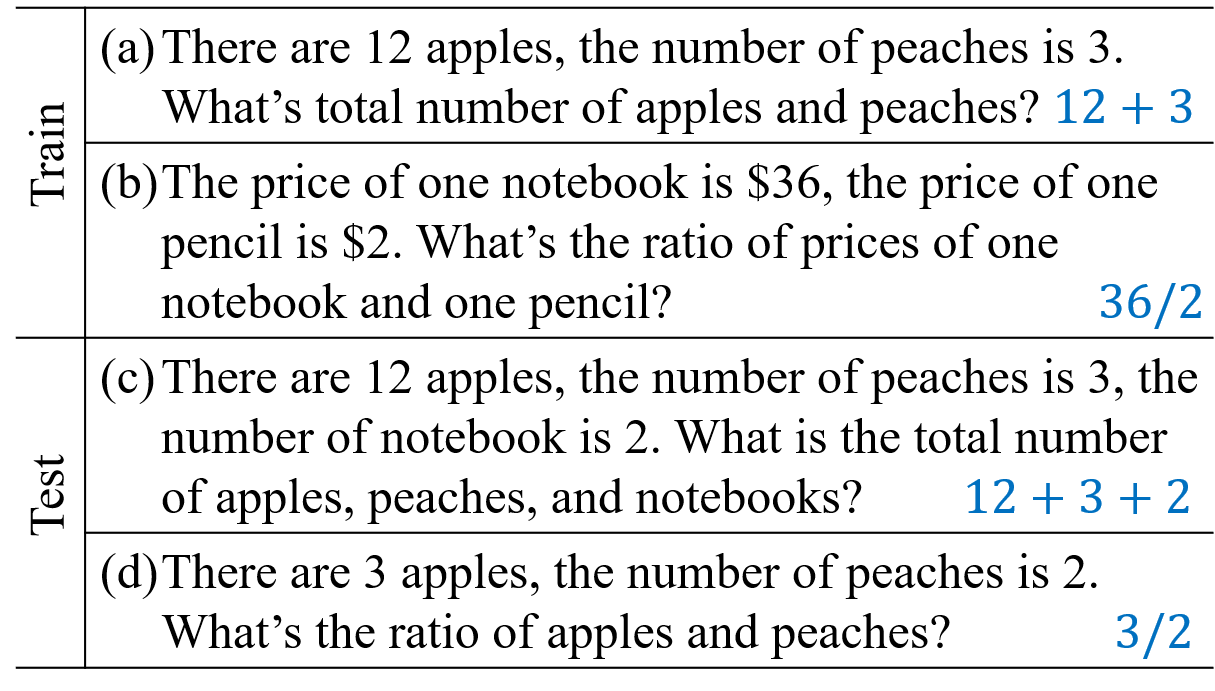}
    \caption{MWP examples with compositional challenges.
    Corresponding expressions are annotated with blue color.}
    \label{fig:motivation}
\end{figure}

\emph{Math Word Problem} (MWP) solving can be formulated as a generation task whose goal is to generate an abstract expression to solve a given MWP.
For example, to solve MWP (a) in Figure~\ref{fig:motivation}, we want to be able to generate the correct expression ``$12 + 3$''.
Among the methods proposed to solve MWPs~\cite{kushman:acl2014, roy:aaai2017, wang:aaai2018, xie:ijcai2019}, many leveraged advanced neural networks and have achieved promising results.
Like many other tasks such as Question Answering~\cite{shaw:arxiv2020,keysers:iclr2020}, MWP solving methods are expected to exhibit \emph{Compositional Generalization}, an important capability to handle novel compositions of known components after learning the ``rules of composition'' from the training data.  
Prior work has shown that models often fail to capture the underlying compositional structure and suffer from big loss of the performance on the data splits with a large compositional gap~\cite{gardner:arxiv2020,kaushik:iclr2020}.

Without exception, an ideal MWP model is expected to correctly infer novel compositions of seen component expressions to answer more new math problems in the text.
Assume \emph{Constituents}, which are the concepts described as background scenario in MWPs (e.g. ``apples'', ``number'', ``price''), are seen in both training and test sets.
If MWP (a) and (b) in Figure~\ref{fig:motivation} are contained in training set, we might encounter following two types of novel \textit{Compositions} during prediction:
(1) Test set contains more complex combinations of expression templates that are not observed.
For example, constituents in (c) have occurred in the training set, but (c) queries about an unseen expression template, which is a compound version of (a).
(2) Test set contains unseen combinations of expression templates and constituents.
In Figure~\ref{fig:motivation}, (d) describes the same scenario as (a) but queries about an expression template that has not occurred in this scenario but occurred in a different scenario like (b) during training.
In the perspective of semantic interpretation, these two types of compositions require structural and lexical generalization of models~\cite{kim:emnlp2020}, respectively.
While it is of utmost importance to investigate compositional generalization on MWP solving, past research has largely overlooked this issue.
There are only few researches research works that studied the out-of-distribution (OOD) phenomenon issues on MWPs ~\cite{hong:aaai2020,kumar:etal2021,patel:naacl2021},
their discussion is limited to length divergence or text perturbation of data, which is only tangentially related to compositional generalization.

This manuscript aims to study compositional generation on MWP solving.
Specifically, we investigate the problem with the following steps:
First, we setup the problems by creating compositional challenges on MWPs.
Next, we propose an effective data augmentation method that could be collaborated with any MWP methods to improve their compositional generalization capabilities.
Last, we examine a set of existing methods as well as our data augmentation method and analyze how the mechanisms make them robust or fragile to the compositional challenges.

In more detail, when creating the compositional challenges, on the one hand, we adopt a data splitting method, which is built upon DBCA~\cite{shaw:arxiv2020} and holds the objective of maximizing the compositional gap with control of constituent difference between training and test.
This method creates desirable compositional splits from large-scale Math23K and MAWPS datasets for evaluation.
On the other hand, we synthesize data via rules to isolate the effect of each type of novel compositions, which results in a synthetic data set with a clear partition of compositions. 
To improve the compositional generalization in MWP solving, inspired by {\small GECA}~\cite{andreas:acl2020good}, we develop a data augmentation protocol aiming to discover the alternative text fragments from MWP corpus and perform a substitution to include more compositional variation into the training data.
Furthermore, the data augmentation procedure is conducted in an iterative manner, where a well-trained MWP solver is leveraged to select the high-qualified augmented data and involve them into MWP corpus for the next round of data augmentation.
During experiments, we compare general encoder-decoder models, strong MWP solving methods, pre-trained models and MWP solving methods combined with our data augmentation method.
We find all tested methods suffer from performance loss when they are exploited under the compositional gap.
By comparison, the structured output decoder is relatively robust.
We also observe common contributions of our data augmentation method on compositional challenges with different effects on different compositions.

To summarize, our contributions include the following aspects:
(1) We introduce two ways to create three compositional MWP datasets to support future research on this topic.
(2) We propose an iterative data augmentation method to improve the compositional generalization capability of general MWP methods.
(3) 
The experiments show that our data augmentation could significantly improve the compositional generalization in MWP solving with the largest increase of $15.6$ percentage points.
(4) 
The evaluation on a wide range of existing methods and further analysis sheds light on what mechanism is potential in bridging the compositional gap.
Our data and code will be publicly available after the acceptance of this manuscript.
\section{Related Work}

\subsection{Math Word Problem Solving}
Early work on math word problem solving task can be categorized into statistical machine learning based methods~\cite{kushman:acl2014,mitra:acl2016,roy:aaai2017,zou:acl2019} and semantic parsing based methods~\cite{shi:emnlp2015,roy:emnlp2015,huang:emnlp2017}. 
The hand-crafted features are collected to represent MWPs and pre-defined templates are leveraged to generate answers.
However, they cannot be applied to large-scale datasets.
Recently, neural networks have been widely applied to solve MWPs.
Seq2seq models were first leveraged to directly transform MWP text sequence to expression sequence~\cite{wang:emnlp2017}, which are still the mainstream MWP solvers until now.
Li \etal~\cite{li:acl2019} borrowed the idea from Transformer and proposed multi-head attention to model different types of features of MWPs.
Many methods managed to encode richer information, they proposed different ways to pre-process MWPs, such as number mapping~\cite{wang:emnlp2017} and graph construction~\cite{zhang:acl2020}.
To capture the structural information of expressions, more advanced work~\cite{Wang:emnlp2018,chiang:naacl2019,liu:emnlp2019,xie:ijcai2019} proposed to decode an expression with implicit or explicit tree structure.
Besides, there are some other studies focusing on knowledge distillation~\cite{zhang:ijcai2020}, weak supervision~\cite{hong:aaai2020} and data augmentation~\cite{liu:arxiv2020} for MWP solving.

\subsection{Compositional Generationation}
Compositional generalization gains much attention from researchers recently. 
People have shown that neural network-based methods always encounter tremendous loss when facing the compositional challenges~\cite{gardner:arxiv2020,kaushik:iclr2020,oren:emnlp2020}.
Different datasets have therefore been introduced to support the compositional generalization research on Language-driven Navigation~\cite{lake:icml2018,ruis:nips2020}, Question Answering~\cite{bahdanau:arxiv2020,keysers:iclr2020,gu:www2021}, Emergent Languages~\cite{chaabouni:acl2020} and text2SQL~\cite{finegan-dollak:acl2018}.
However, there is a lack of study of compositional generalization on MWP solving task.
Hong \etal~\cite{hong:aaai2020} briefly discussed the OOD challenge by splitting the MWP data based on the length of MWPs.
Kumar \etal~\cite{kumar:etal2021} discussed the robustness of MWP solvers under the adversarial attack of question reordering and sentence paragraphing.
Patel \etal~\cite{patel:naacl2021} included carefully chosen text variations to MWPs for more robust evaluation of methods.
The focus of them is more about length divergence and text perturbation rather than compositional generalization.
\section{Problem Setup}
\label{sec:setup}

A MWP is a natural language narrative associated with a question towards the narrative whose answer is a numerical value. 
It consists of multiple sentences $q = \{s_1, ..., s_i, ..., s_k\}$ where $s_i$ is a sequence of tokens $\{w_1, ..., w_m\}$. 
Each token could be either a word token or a quantity.
The goal of MWP solving is to generate an expression $e$ based on $q$, which is formed by quantities and pre-defined mathematical operators and which, when evaluated, gives the numerical answer to $q$.
Assume we have a set of training data $\mathcal{D}^p = \{(q^p_1, e^p_1), ..., (q^p_{|\mathcal{D}^p|}, e^p_{|\mathcal{D}^p|})\}$, where each data point is a MWP-expression pair.
We are supposed to leverage $\mathcal{D}^p$ to train a MWP model that can make prediction on a test dataset $\mathcal{D}^q$.
We define \emph{a challenging dataset for evaluating compositional generalization to be a dataset where there is a potentially large compositional difference but small constituent difference between  $\mathcal{D}^p$ and $\mathcal{D}^q$}.
And we expect a good MWP model to have the capability to maximize the performance on $\mathcal{D}^q$ despite the compositional gap of $\mathcal{D}^p$.

To investigate the compositional generalization in MWP solving, we need to first construct challenging datasets as described above.
To this end, we present a data splitting method and a data synthesizing method, which aim at generating large-scale compositional splits and isolating compositional effect, respectively.

\subsection{Split Data with DBCA}
\label{sec:split_method}

As introduced in the Sec.~\ref{sec:intro}, existing work~\cite{keysers:iclr2020,shaw:arxiv2020} has proposed their method based on distribution-based compositional assessment (DBCA) to create compositional splits for KBQA task, which could be summarized as follows:
\begin{itemize}
    \item Each data is represented using a graph, where the nodes are considered as \textit{Atoms} and the rule applications on the graph are treated as \textit{Compounds}.
    \item A distribution-based compositional assessment~\cite{keysers:iclr2020} is utilized to measure the divergence:
    $\text{DBCA}(\mathcal{D}^p, \mathcal{D}^q) = 1 - \mathcal{C}_{\alpha}(\mathcal{P} \lVert \mathcal{Q})$,
    where $\mathcal{C}_\alpha(\mathcal{P} \lVert \mathcal{Q}) = \sum_i p^\alpha_i q^{1-\alpha}_i \in [0, 1]$ is the Chernoff coefficient~\cite{chung:jmaa1989}, $\mathcal{P}$ and $\mathcal{Q}$ are distributions deriving from training and test sets, respectively.
    In more detail, they obtain atom and compound distributions based on their frequencies, and then compute \textit{Atom Divergence} $\text{DBCA}^a$ and \textit{Compound Divergence} $\text{DBCA}^c$, respectively.
    As we can see, the atom divergence measures the constituent difference and compound divergence measures the compositional difference of a dataset. 
    \item Starting from a data pool $\mathcal{D}$, they apply a greedy algorithm to assign each data to form $\mathcal{D}^p$ and $\mathcal{D}^q$, the objective of which is to maximize $\text{DBCA}^c$ with the control of an upper bound of $\text{DBCA}^a$.
\end{itemize}
The detailed algorithm could be found in original work~\cite{shaw:arxiv2020}.

\begin{figure}[!t]
    \centering
    \includegraphics[width=0.48\textwidth]{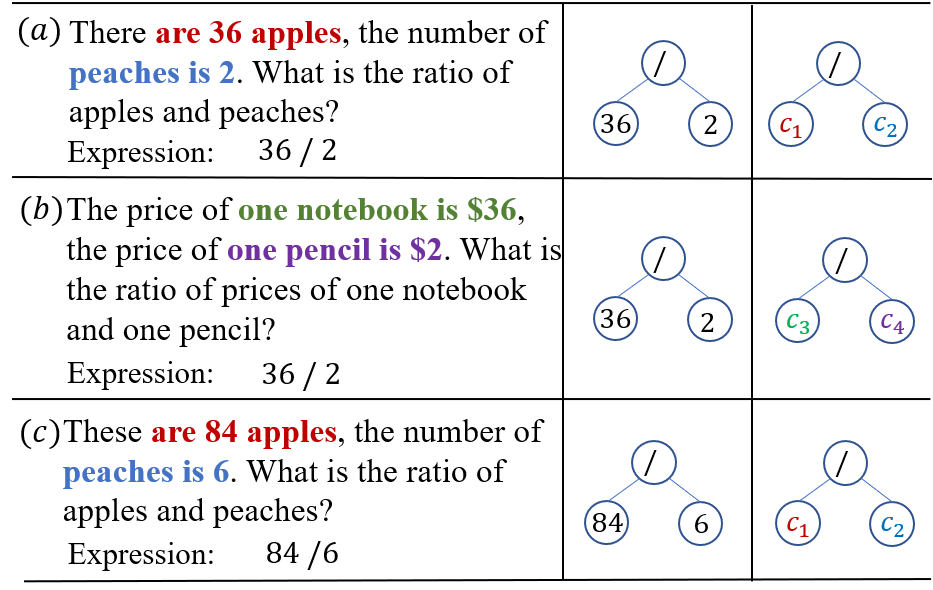}
    \caption{Three examples of data. 
    MWPs and expressions are shown in the left column.
    The original expression trees and our trees are shown in the middle and right columns, respectively.
    The cluster indexes and their corresponding contexts are shown in the same colors.
    }
    \label{fig:tree}
\end{figure}

When it comes to MWPs, each expression of MWPs could be represented using an expression tree~\cite{zhang:acl2020}, which can be treated as a specific graph.
The leaf nodes are quantities and they are connected via mathematical operators.
However, the lack of semantics of the quantities leads to inapplicability of the above divergence measurement.
As shown in Figure~\ref{fig:tree}, (a) and (b) describe the different scenarios but their graphs are exactly the same.
(a) and (c) describe the similair scenarios but with different quantities, which leads to different graphs. 
Therefore, we enrich the representation of quantity in the expression tree by its contexts.
Specifically, we extract tokens within a window slide centered on a target quantity in a MWP to represent the quantity.
Then, we represent the quantities via TF-IDF vectors and perform $k$-means clustering algorithm based on the vector-based representations, which assigns a cluster index to each quantity\footnote{We have also tried other methods such as Word2vec, BERT to represent the context. They cannot outperform TF-IDF vectors even though they have more expressive architectures.
It could be explained that TF-IDF vectors are derived from the MWPs domain so it contains more domain knowledge while other models involve more general knowledge.}.
As a result, the quantities within similar contexts are labeled with the same cluster indexes and we treat such quantities annotated with cluster indexes as atoms.
In Figure~\ref{fig:tree}, after representing the trees featured with context information, (a) and (c) will have the same graphs.
Next, $\text{DBCA}^a$ is derived from the frequency of cluster indexes after traversing the expression tree.
To obtain $\text{DBCA}^c$, we exhaustively search sub-expression trees in a complete expression tree.
Each sub-expression tree consists of a left component, an operator node and a right component, where a component is either a sub-expression tree or an atom in the expression tree.
Afterwards, we follow the traditional distribution-based assessment and the greedy algorithm to generate realistic compositional splits. 

\subsection{Synthesize Data to Isolate Effect}
\label{sec:synthesize_method}

\begin{figure*}[!t]
    \centering
    \includegraphics[width=0.9\textwidth]{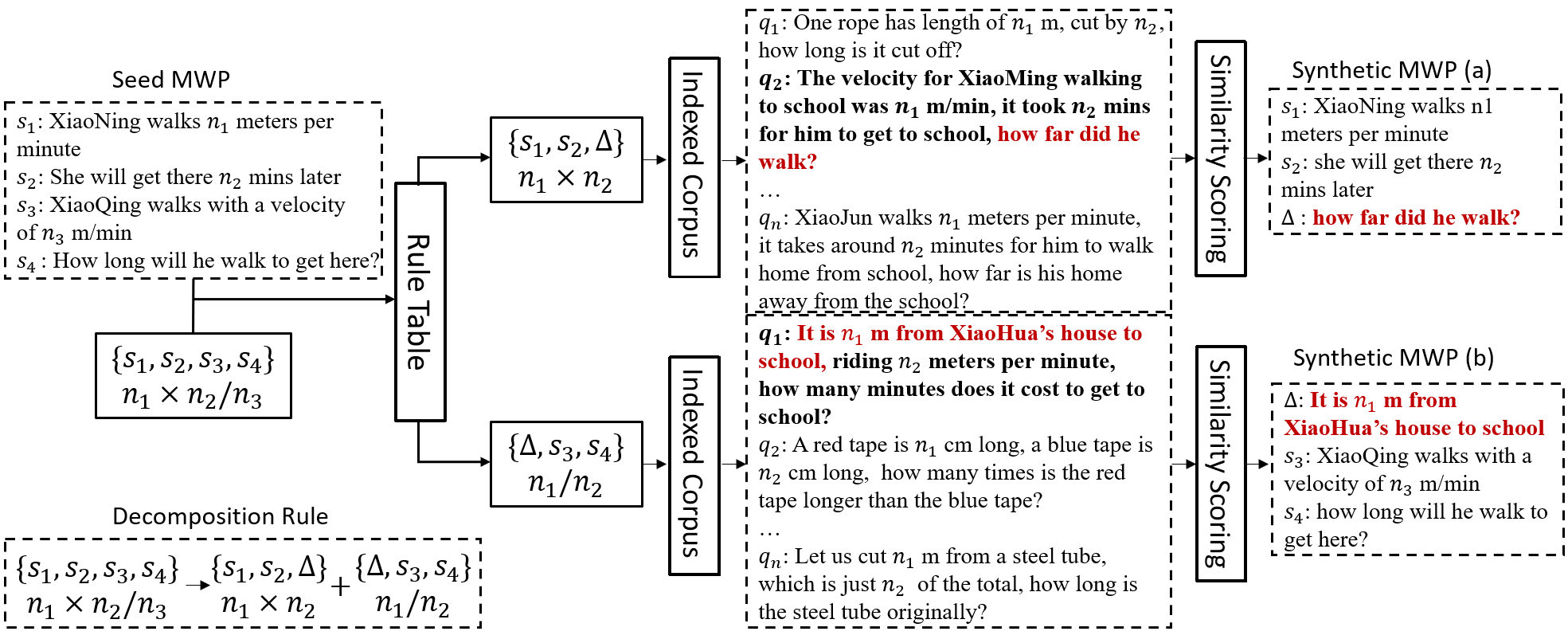}
    \caption{
    One example of generating synthetic MWPs with a seed MWP via a decomposition rule.
    The selected MWP in the corpus is shown in bold line and the instantiate sentence of $\Delta$ is annotated with red color.
    }
    \label{fig:synthetic_data}
\end{figure*}

Data splitting method results in datasets differing in compositions with a fuzzy boundary.
Since our goal is to understand how MWP solvers respond to different compositions precisely, to disentangle compositional gaps arising from new expression templates or new combinations of templates and constituents, we synthetically construct data via rules.

As the preliminary step, we pre-define a \textit{Rule Table}~\footnote{The complete table and more implementation details are displayed in Appendix}, which contains the rules that we follow to synthesize data.
Each rule states that given a MWP with an expression template, it could be transformed into new MWP(s) with certain expression template(s).
For example, the MWP (d) in Figure~\ref{fig:motivation} could be represented by $\{s_1, s_2, s_3, s_4\}$ with expression template $n_1 \times n_2 / n_3$, where the quantity in the expression is replaced by the position index~\cite{wang:aaai2018}.
Following one rule, we could decompose it into $\{s_1, s_2, \Delta, s_4\}$ with template $n_1 \times n_2$ and $\{\Delta, s_3, s_4\}$ with template $n_1 / n_2$, where $\Delta$ indicates a placeholder to be instantiated.
We further generate an \textit{Indexed Corpus} by indexing a set of initial MWPs with their templates as keys and MWPs as values, which could be utilized to instantiate the placeholders.
Moreover, we define a \textit{Similarity Scoring} function to measure the semantic similarity between MWPs.
Specifically, we extract a collection of nouns, verbs and measure units from the MWPs using existing toolkits.
When we measure the similarity of two MWPs, we compare the token-level overlap.
A large similarity value indicates the MWPs share similar constituents. 
Next, we introduce how we synthesize data with the two types of compositions:

\begin{itemize}
    \item \textbf{Decomposition}. We treat each decomposable $q \in \mathcal{D}$ as the seed MWP, where $\mathcal{D}$ is a data pool.
    We employ applicable decomposition rules from the rule table to generate two uninstantiated MPWs.
    We retrieve all MWPs from the indexed corpus with their corresponding templates, select the MWP which has the highest similarity score to the seed MWP and instantiate $\Delta$ with corresponding sentence of the selected MWP.
    This step will generate two synthetic MWPs $\{q^{'}, q^{''}\}$ specific to $q$.
    To illustrate the above procedure better, we display an example in Figure~\ref{fig:synthetic_data}.
    Eventually, we allocate $(q, e)$ into test set and $\{(q^{'}, e^{'}), (q^{''}, e^{''})\}$ into training set to form $\mathcal{D}^q$ and $\mathcal{D}^p$, respectively.
    It is worth noting that sometimes the selected MWP $q^{'}$ or $q^{''}$ has similarity score value to $q$ as $0$, in this case, we abandon it. 
    This is to ensure the similar constituent distribution between $\mathcal{D}^p$ and $\mathcal{D}^q$.
    \item \textbf{Reformulation}. For each applicable $q \in \mathcal{D}$, we employ reformulation rules to generate one uninstantiated MWP and instantiate placeholder $\Delta$ as illustrated as above to obtain a synthetic MWP $q'$.
    If for any existing $q'' \in \mathcal{D}^p$ which has $e''=e'$ and the similarity score to $q'$ is $0$, we include $(q, e)$ into $\mathcal{D}^p$ and $(q', e')$ into $\mathcal{D}^q$. Otherwise, we abandon it.
    This is because we would like to only include the MWP of which the combination of templates and constituents have not been seen during training into the test set.
\end{itemize} 

After above procedure, we obtain the synthetic data sets $\mathcal{D}^p$ for training and $\mathcal{D}^q$ for testing.
$\mathcal{D}^p$ includes all the MWPs having simple templates as their expressions, $\mathcal{D}^q$ contains all the MWPs which are either new compounds of the simple  expression templates or new combinations of existing constituents and simple expression templates. 
As synthesized MWPs always share constituents with seed MWP, the constituent difference between training and test sets is well-controlled and the only variation is the compositionality.


\section{Data Augmentation for MWP Solvers}
\label{sec:data_aug}

\begin{figure*}[!t]
    \centering
    \includegraphics[width=\textwidth]{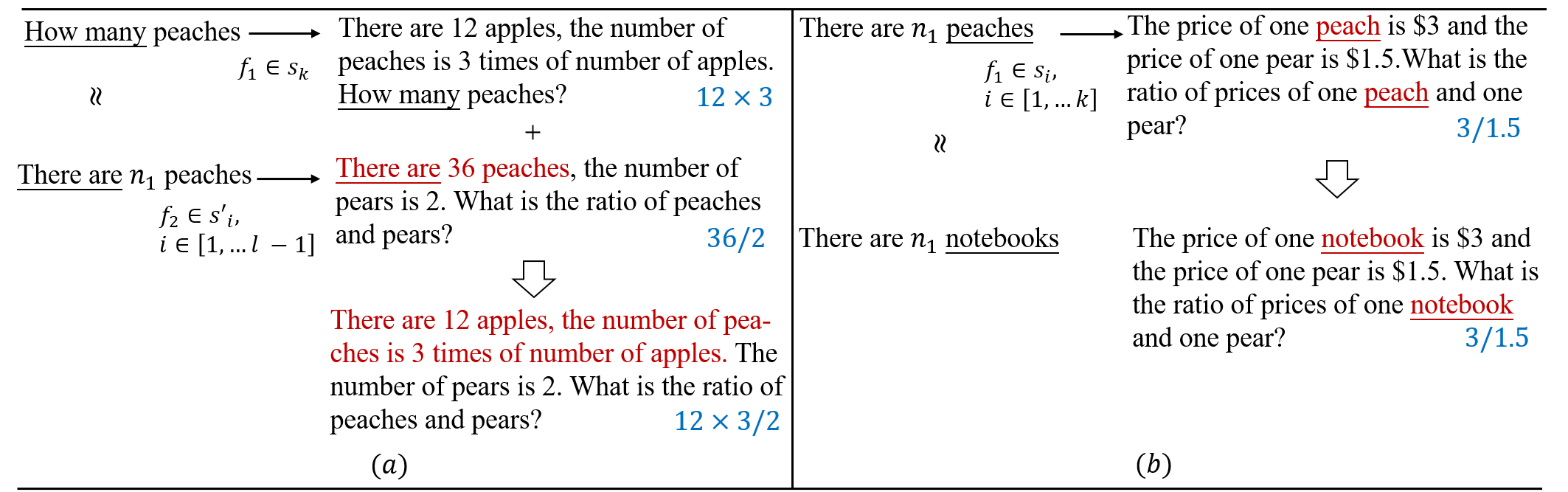}
    \caption{Two examples of data augmentation protocol on MWPs via operation 1 and operation 2 in sub-figure (a) and (b), respectively. 
    Fragments are labeled with underline and substituent parts in the synthetic MWPs are in red color.
    Expressions are annotated with blue color.
    }
    \label{fig:cg_method}
\end{figure*}

Previous studies~\cite{jiang:acl2016, andreas:acl2020good} have conducted discussions on data augmentation, which has proven to be effective in improving the robustness of models.
In this manuscript, we propose an iterative data augmentation method that aims to generate more compositional variation without involving external data and human annotation. 
Inspired by {\small GECA}~\cite{andreas:acl2020good}, we discover the alternative text fragments and substituent the text fragments in existing MWPs to synthesize new MWPs.
We further improve the quality of the generated data by measuring the confidence of new MWPs via a well-trained MWP solver.
The entire procedure is displayed in Algorithm~\ref{alg:data_augment}.
As we can see, this algorithm is designed with an iterative framework, where a \textit{Data Augmentation Protocol} takes charge of generating new MWPs and a \textit{Data Ranker} plays the role of judging the quality of the generated MWPs.
After $T$ rounds of iteration, the augmented data will fill the compositional gap which is leveraged to train a robust MWP model.

\begin{algorithm}[H]
\SetAlgoLined
Input: initial training data set $\mathcal{D}^p$, max iteration $T$
\\
\For{i in T}{
$\mathcal{D}^*$ = $\emptyset$ \\
\For{q in $\mathcal{D}^p$}{
$\mathcal{D}_q^p$ = \textbf{DataAugmentationProtocol}($q$) \\
$\mathcal{D}^* = \mathcal{D}^* \cup \mathcal{D}_q^p$
}
$\mathcal{D}^*$ = \textbf{DataRanker}($\mathcal{D}^*$) \\
$\mathcal{D}^p$ = $\mathcal{D}^p \cup \mathcal{D}^*$ 
}
Return: augmented training data set $\mathcal{D}^p$

\caption{Data Augmentation Algorithm}
\label{alg:data_augment}
\end{algorithm}

We first describe our data augmentation protocol which is an improved version of {\small GECA}.
Two text fragment is defined as \textit{Alternative} if they occur in some common environments.
With the alternative fragments, we define the following operations to involve new compositions in a MWP:
(1) If the first fragment is from the last sentence of a MWP, which is the question sentence, and the second fragment is not from the last sentence of another MWP, we use the sentences excluding the last sentence of the first MWP to substituent the corresponding sentence of the second MWP.
In the example of Figure~\ref{fig:cg_method} (a), ``how many'' and ``there are'' are alternative fragments as they appear in the same environment ``peaches''.
Given a MWP querying ``how many'', we could leverage its narrative description to substituent the sentence ``there are 36 peaches'' in another MWP to form a new MWP.
(2) If both the fragments are not from the last sentences of MWPs, we replace all the first fragment in a MWP with the second fragment.
In the example of Figure~\ref{fig:cg_method} (b), ``peach'' and ``notebook'' is a pair of alternative fragments.
When there is a MWP with fragment ``peach'', we can generate a new MWP by replacing all the ``peach'' with ``notebook''.
As we can see, the first operation involves more diversity to the expression structure and generates MWPs with novel expression templates.
The second operation helps to generate MWPs with novel combinations of expression templates and constituents.

We now formally describe the above operations.
Recall each MWP consists of multiple sentences $q=\{s_1, ..., s_k\}$. 
A fragment is a set of non-overlapping spans of $s$.
An environment is $s$ with a text fragment removed.
We denote fragment and environment as $f$ and $s/f$, respectively.
If there are environments $x=s_1/f_1$ and $y=s_2/f_2$ with $x=y$, $(f_1, f_2)$ is a pair of alternative fragments:

\begin{itemize}
\item \textbf{Operation 1}.
Given $q=\{s_1, ..., s_k\}$ and $q'=\{s'_1, ..., s'_l\}$ with $e$ and $e'$, respectively.
If $f_1 \in s_k$ and $f_2 \in s'_i$, where $i \in [1, ..., l-1]$, we could form a new MWP $q'=\{s'_1, ..., s'_{i-1}, s_1, ..., s_{k-1}, s'_{i+1}, ..., s'_l\}$. 
The expression of this MWP is $e'$ modified by replacing the quantity in $s'_i$ with $e$.
\item \textbf{Operation 2}.
Given $q=\{s_1, ..., s_k\}$ with $e$, if $f_1 \in s_i$ where $i \in [1, ..., k-1]$, we could form a MWP $q'=\{s'_1, .., s'_i, ..., s'_k\}$ where $s'_i$ is obtained via replacing all the $f_1$ in $s_i$ with $f_2$. 
The expression of this MWP is still $e$.
\end{itemize}
For each $q \in \mathcal{D}^p$, we generate a set of new MWPs, via either operation 1 or operation 2.
These newly generated MWPs will form a set $\mathcal{D}^*$.

As the data augmentation protocol follows simple patterns, incorrect fragment discovery and illogical fragment substitution would be easily involved and deemed as noisy data.
To improve the quality of the augmented data, we leverage a MWP solver to perform as the data ranker to filter out the sub-optimal augmented data.
We assume that a well-trained MWP solver could learn the capability of understanding general narrative of MWPs.
Ambiguous and illogical description would be considered as uncertain data by a well-trained MWP solver, which results in the generation of an expression with low confidence~\cite{adebayo:nips2020,antoran:iclr2021}. 
Therefore, for $q \in \mathcal{D}^*$, we employ a well-trained MWP solver to obtain its expression with a sequence of probabilities and sort them based on the accumulative product of the probabilities.
The MWPs with low probabilities are filtered out from $\mathcal{D}^*$ and the rest are merged with the initial training data set which are either used to generate more data for the next iteration or returned as the augmented data set.
It is worth noting that there is no data leakage happening as the test data is hidden during the entire augmentation procedure.

\textbf{Discussion:} Compared with the original protocol proposed by Andreas \etal~\cite{andreas:acl2020good}, we improve their algorithm by developing novel synthetic rules to fit the nature of compositional MWPs and collaborating with an iterative framework to continuously improve the quality of the augmented data.
Compared with the synthesizing procedure described in Sec.~\ref{sec:synthesize_method} which heavily relies on manual rules of expression annotations, this data augmentation method generates data based on the patterns of MWPs and does not depend on any hand-crafted rules.

\section{Experiments}
\label{sec:exp}

In this section, we first introduce the compositional data sets that we created in Sec.~\ref{sec:datasets}.
Next, we describe our experimental setup in  Sec.~\ref{sec:baselines}.
Then, we show the evaluation of different methods on compositional challenges and analysis in Sec.~\ref{sec:results} and Sec.~\ref{sec:analysis}, respectively.

\subsection{Compositional Data Sets}
\label{sec:datasets}

We create compositional challenges deriving from two datasets: Math23K dataset from Wang \etal~\cite{wang:emnlp2017} and MAWPS dataset from Kedziorski \etal~\cite{koncelkedziorski:tacl2015}.
These two datasets are both large-scale MWP datasets that are commonly used in existing work.
We utilize NLTK\footnote{\url{https://www.nltk.org/}} as the toolkit to extract contextual features through tokenization, POS tagging and dependency parsing.
For data splitting, scikit-learn tool\footnote{\url{https://scikit-learn.org/stable/modules/generated/sklearn.feature_extraction.text.TfidfVectorizer.html}} is employed to implement TF-IDF.
We choose $k$ value from range $\{5, 10, 20\}$ based on the highest $\text{DBCA}^c$ it could achieve after splitting. 
Following previous implementation\footnote{\url{https://github.com/google-research/language/tree/master/language/nqg}}, $\alpha$ in DBCA is set as $0.1$ and $0.5$ for $\text{DBCA}^a$ and $\text{DBCA}^c$, respectively.
As a result, we obtain the following three compositional data sets:

\begin{table}[!t]
\begin{center}
\small
\begin{tabular}{>{\centering\arraybackslash}p{1.3cm} |>{\centering\arraybackslash}p{0.9cm}|>{\centering\arraybackslash}p{1.3cm} >{\centering\arraybackslash}p{0.2cm} >{\centering\arraybackslash}p{0.7cm}>{\centering\arraybackslash}p{0.9cm}}
\toprule
& \textbf{Split} & \textbf{NR} & \textbf{LR} & \textbf{$\text{DBCA}^a$} & \textbf{$\text{DBCA}^c$} \\
\midrule
\multirow{2}{*}{Math23K} & I.I.D. & 18543/4619 & 0.99 & 0.00 & 0.20 \\ 
 & ComDiv & 18543/4619 & 1.01 & 0.00 & 0.69 \\ 
 \hline
\multirow{2}{*}{MAWPS} & I.I.D. & 1899/474 & 1.01 &0.00  & 0.09 \\
 & ComDiv & 1899/474 & 0.91 & 0.01 & 0.77 \\
 \hline
SD & ComDiv & 1078/745 & 0.76 & 0.01 & 0.82 \\
\bottomrule
\end{tabular}
\caption{Statistics of datasets. 
\textbf{NR}, and \textbf{LR} denote Number Ratio and text Length Ratio between training and test sets, respectively.
\textbf{$\text{DBCA}^a$} and \textbf{$\text{DBCA}^c$} are atom and compound divergences, respectively.
}
\label{tab:datasets}
\end{center}
\end{table}

\begin{table*}[!t]
\begin{center}
\small
\scalebox{0.88}{
\begin{tabular}{l|ccc|ccc|ccc}
\toprule
 & \multicolumn{3}{c|}{{\bf{Math23K}}} &\multicolumn{3}{c}{{\bf{MAWPS}}} & \multicolumn{3}{c}{{\bf{SD}}} \\ 
\textbf{Methods} & \textbf{I.I.D.} & \textbf{ComDiv} & \textbf{Rel. Gap} & \textbf{I.I.D.}  & \textbf{ComDiv} & \textbf{Rel. Gap} & \textbf{Valid} & \textbf{Test} & \textbf{Rel. Gap}\\
\midrule
LSTM  &  67.4 &  47.4 
& 0.42 
& 74.5  & 36.5 & 1.04 
& 93.7 & 7.8 & 11.0 \\ 
Transformer  &  58.0 &   40.3 & 0.44
& 72.6 & 40.1 & 0.81 
& 91.0 &6.6 & 12.8 \\
\midrule
BERTGen &  68.1 &  43.0
& 0.58
& 70.7 & 43.0 & 0.64 
& 96.9 & 9.6 &  9.09\\
GPT-2  &  70.4 &  49.7 &  0.42
& 62.2  & 31.9  & 0.95
& 95.1 & 8.3 & 10.5\\
\midrule
MathEN  &  66.5 &  46.9 &  0.42 
& 76.4  & 40.9 & 0.87 
& 94.6 & 7.5 & 11.7\\
GTS  &  71.5 &  51.2 & 0.39 
& 76.9 & 48.4 & 0.59
& \textbf{96.9} & 7.9 &11.3  \\
\midrule

MathEN+$\mathrm{DA}$
&67.3{\color{red}(+0.8)} & 47.5{\color{red}(+0.6)} & 0.42 & 
 77.6 {\color{red}(+1.2)}& 42.4{\color{red}(+1.5)} & 0.83
& 95.5 {\color{red}(+0.9)}& 17.4{\color{red}(+9.9)} & 4.5 \\

GTS+$\mathrm{DA}$
& \textbf{72.1}{\color{red}(+0.6)} & \textbf{52.0}{\color{red}(+0.8)} &\textbf{ 0.38} & 
\textbf{78.1}{\color{red}(+1.2)} & \textbf{49.8}{\color{red}(+1.4)} &  \textbf{0.57}
&96.4{\color{red}(-0.5)} & \textbf{23.5} {\color{red}(+15.6)}& \textbf{3.1}\\

\bottomrule
\end{tabular}
}
\caption
{Answer accuracy of four groups of methods on different data sets.
We show the accuracy of test data for different splits of Math23K and MAWPS, the accuracy of validation and test data for SD data set.
To show the performance gap between normal and compositional splits, we compare the results of Math23K and MAWPS with I.I.D. and ComDiv splits, results of SD with validation and test sets, respectively.
\textbf{Rel. Gap} denotes $\text{Acc}_{\text{I.I.D.}}/\text{Acc}_{\text{ComDiv}}-1$.
The numbers in red font denote improvement brought by data augmentation.
}. 
\label{tab:main_result}
\end{center}
\end{table*}

\begin{itemize}[leftmargin=*]
    \item \textbf{Math23K-ComDiv:} This is a dataset created by performing data splitting as described in Sec.~~\ref{sec:split_method} on Math23K.
    It contains Chinese MWPs associated with corresponding expressions.
    \item \textbf{MAWPS-ComDiv:} We conduct data splitting as described in Sec.~\ref{sec:split_method} on MAWPS and obtain this compositional splits. 
    It contains English MWPs which query a variety of arithmetic math expressions.
    \item \textbf{SD:} This is a synthetic dataset created by conducting data synthesis as described in Sec.~\ref{sec:synthesize_method} with Math23K performing as its data pool.
    After the automatic synthesis procedure, we further do manual proofreading to ensure the validity of this data set.
\end{itemize}

We employ above compositional challenges to evaluate compositional generation of a series of methods.
As a comparison, we also include original \textbf{I.I.D.} splits of Math23K and MAWPS.
Table~\ref{tab:datasets} displays the statistics of all datasets.
As we can see, most datasets have LR value approximately to $1$, which indicates that the questions with different length are evenly distributed over training and test sets.
The compositional data sets all have high $\text{DBCA}^c$ but low $\text{DBCA}^a$, which indicates the datasets indeed follow our expected objective of maximizing the compound divergence with great control of small constituent divergence.

\subsection{Experimental Setup}
\label{sec:baselines}

Following the standard training and prediction diagrams of MWP solving, we treat it as a generation task with the supervision of ground truth expressions.
During prediction, we report the answer accuracy by comparing the execution results of generated expressions and the correct answers.

\paragraph{Comparable Methods.}

The methods tested on compositional challenges of MWPs are mainly categorized into four families: %
(1) General encoder-decoder models, i.e., \textbf{LSTM}~\cite{vaswani:nips2017} and \textbf{Transformer}~\cite{hochreiter:nc1997}.
(2) Strong MWP solving methods
, i.e., \textbf{MathEN}~\cite{Wang:emnlp2018} and \textbf{GTS}~\cite{xie:ijcai2019}.
(3) Pretrained models i.e., \textbf{BERTGen}~\cite{chen:acl2020} and \textbf{GPT-2}~\cite{radford:2019}, which have been suggested to improve general compositional generalization~\cite{oren:emnlp2020}.
(4) MathEN and GTS collaborating with our proposed data augmentation method (refer to as \textbf{MathEN+DA} and \textbf{GTS+DA}), which aims to improve compositional generalization.

\paragraph{Implementation Details.}
\label{sec:implementation}

All the experiments are implemented by PyTorch on Nvidia V440.64.00-32GB GPU cards.
We implement the existing work by leveraging their official code or following their methodology description.
Regarding general encoder-decoder models, we set the same embedding size and hidden size as MathEN. 
Regarding pretrained baselines, we adopt base BERT and base GPT-2. 
Regarding MWP baselines and MWP baselines with data augmentation, we keep hyperparameters the same as original manuscripts\footnote{We show detailed hyperparameters in Appendix.}.
For each type of data splits, we randomly allocate 20\% data from training to do validation.
In terms of the data augmentation method, the MWP solvers well trained with the augmented data in the last round of iteration are used to rank the data.
We set the maximum iteration time $T$ as 7 and keep the top $20\%$ augmented MWPs for each iteration.

\subsection{Results}
\label{sec:results}

Table~\ref{tab:main_result} displays the results of all tested methods.
Based on the table, we have the following observations:
(1) For SD data set, we observe a huge performance gap between validation set and test set. 
A similar performance gap appears between I.I.D. split and ComDiv split on both Math23K and MAWPS.
This indicates that all families of methods are vulnerable to compositional challenges MWP solving.
Among these three datasets with compositional challenges, SD dataset causes the biggest loss. 
This could be explained by the nature of big compound divergence of SD data.
(2) Comparing the first three sections, in terms of the most robust method to compositional challenges, there is no absolute agreement on all the data sets.
Nevertheless, GTS shows relatively good compositional generation as it has the lowest Rel.Gap on Math23K and MAWPS. 
This may because the tree-structured neural module which outputs an expression tree is more powerful when generalizing to unseen compositions.
(3) Our data augmentation could provide positive effect to MathEN and GTS methods. 
This indicates that data augmentation is effective in improving the compositional generalization capability of general MWP methods.
Meanwhile, data augmentation shows the significant contribution on SD dataset with $9.9$ and $15.6$ percentage points improvement to MathEN and GTS, respectively.
This implies it indeed plays a key role in bridging the gap of compositions.
It's worth noting that data augmentation provides the largest performance gain 15.6\% to GTS model on SD test set.
It may because that data augmentation includes more lexical and structural variations for training and tree-structured neural network has a better inductive bias to capture them, which is also verified in prior work~\cite{mccoy:acl2019}.

\subsection{Further Analysis}
\label{sec:analysis}

\ignore{
\begin{figure}
\centering
    \begin{subfigure}{\linewidth}          
        \includegraphics[scale=0.25]{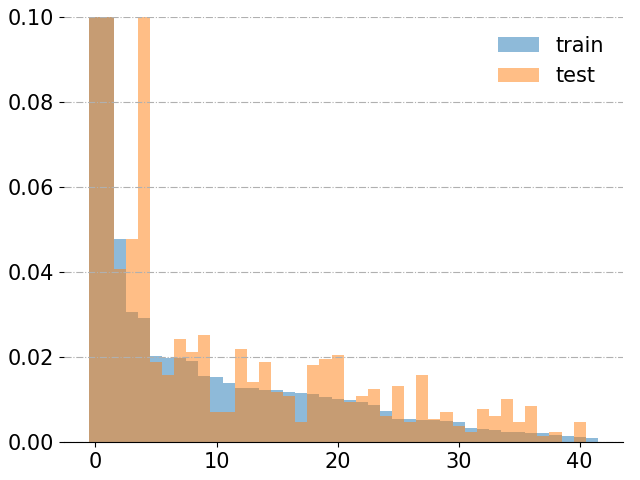}
        \caption{}
    \end{subfigure}%

    \begin{subfigure}{\linewidth}
        \includegraphics[scale=0.25]{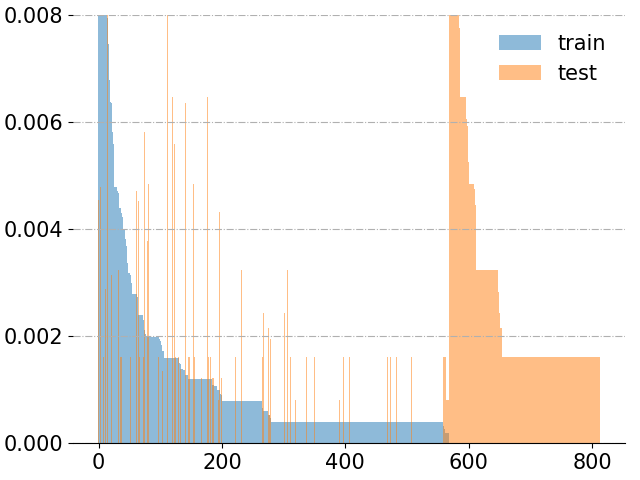}
        \caption{}
    \end{subfigure}
\caption{Distributions of (a) atoms and (b) compounds in the training vs. test set of our splitting method.}
\label{fig:frequency}
\end{figure}

\paragraph{Atom and Compound Distributions}
To verify that our splitting method could effectively retain the objective of maximizing $\text{DBCA}^c$ with a upper bound of $\text{DBCA}^a$, we display Figure~\ref{fig:frequency} to show the distribution of atoms and compounds in the training and test sets when $k=40$ for better visualization.
As we can see, the atom frequency of these two sets are similar while the compounds are distributed differently.
}

\paragraph{Accuracy with Increasing Compound Divergence}


\begin{figure}
\centering
    \subfigure[]{\includegraphics[width=0.26\textwidth, trim=0 1cm 0 0, clip]{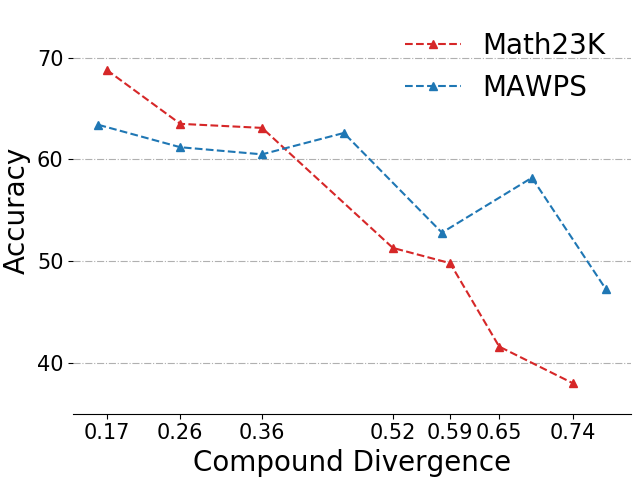}}
    \hspace{-0.5cm}
    \subfigure[]{\includegraphics[width=0.24\textwidth, trim=0.3cm 1cm 0 0, clip]{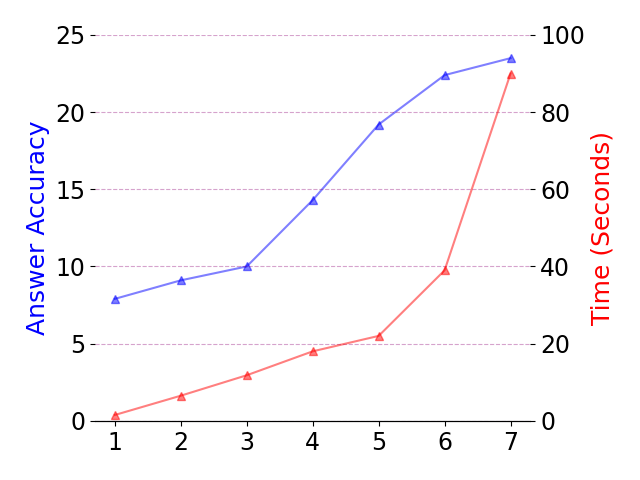}}
\caption{(a) Performance changes of MathEN with the increase of $\text{DBCA}^c$ on Math23K and MAWPS. (b) Evolution of answer accuracy (blue) and average training time per epoch (red) of GTS on SD dataset with the increase of iteration time.}
\label{fig:compound_divergence}
\end{figure}

To show the influence of compound divergence described in Sec.~\ref{sec:split_method}, we display Figure~\ref{fig:compound_divergence} (a).
As we can see, the performance of MathEN on both Math23K and MAWPS datasets decreases dramatically with the increase of compound divergence $\text{DBCA}^c$. 
This indicates the performance is sensitive to the change of compound divergence, which verifies that the compound divergence is indeed an effective indicator for measuring the compositional challenge of MWPs.
Therefore, it is reasonable for us to split MWPs data and obtain the compositional data sets with $\text{DBCA}^c$ performing as the measurement criteria.

\paragraph{Effect of Different Composition Types} 

\begin{table}[t!]
\centering 
\small
\begin{tabular}{l |ccc}
\toprule
& \multicolumn{2}{c}{{\bf{SD}}} \\
 & \textbf{Structural} &  \textbf{Lexical} \\
\midrule
MathEN & 3.0 & 46.3 \\
GTS & 2.9 & 51.3 \\
\midrule
MathEN+DA & 14.2 & 47.5 \\
GTS+DA & 16.8 & 63.8 \\
\bottomrule
\end{tabular}  
\caption{Results on partition of SD dataset with Structural and Lexical generalization separately.
}
\label{tab:diff_comp}
\end{table}

SD dataset that we introduced in Sec.~\ref{sec:synthesize_method} is created for isolating the effect of different types of compositions. 
Decomposition and reformulation rules create data requiring structural and lexical generalization, respectively.
We display Table~\ref{tab:diff_comp} to discuss the effect of different rules.
From the table, we observe that new compound expression templates bring much more difficulties to models than new combinations of expression templates and constituents.
It is easy to understand as template inferring requires a deep understanding towards the expression.
We also observe that data augmentation benefits more to MWPs with new compound expression templates.

\ignore{
\paragraph{Effect of Different Rules in Data Augmentation} 

\begin{table}[t!]
\centering 
\small
\begin{tabular}{l |ccc}
\toprule
& \textbf{MAWPS} & \textbf{SD} \\
\midrule 
MathEN+DA & xxx & xxx   \\
MathEN+DA (Composition) & xxx & xxx   \\
MathEN+DA (Re) & xxx & xxx \\
\bottomrule
\end{tabular}   
\caption{Improvement via different rules of data augmentation on MAWPS and Synthetic Data (\textbf{SD}). 
}
\label{tab:data_aug}
\end{table}

\begin{table}[t!]
\centering 
\small
\begin{tabular}{l |ccc}
\toprule
& \textbf{SD (Composition)} & \textbf{SD} \\
\midrule 
MathEN+DA & xxx & xxx   \\
MathEN+DA (Rule 1) & xxx & xxx   \\
MathEN+DA (Rule 2) & xxx & xxx \\
\bottomrule
\end{tabular}   
\caption{Improvement via different rules of data augmentation on MAWPS and Synthetic Data (\textbf{SD}). 
}
\label{tab:data_aug}
\end{table}

We analyze the contribution of rule 1 and rule 2 in data augmentation in Table~\ref{tab:data_aug}.
As we can see, their contribution varies for different datasets. 
As the two rules serve different composition types, the datasets with more novel compound expressions will benefit more from rule 1.
Same way to rule 2.
}

\paragraph{Ablation Study of Data Augmentation}

\begin{table}[t!]
\centering 
\small
\begin{tabular}{l |cc}
\toprule
& \textbf{Math23K-ComDiv} & \textbf{MAWPS-ComDiv} \\
\midrule
MathEn+DA & 47.5 & 42.4  \\
- DataRanker & 46.6 & 41.4 \\
- Operation 1 & 47.3 & 41.8 \\
- Operation 2 & 47.2 & 41.6 \\
\bottomrule
\end{tabular}  
\caption{Ablation study of data augmentation with MathEn on Math23K-ComDiv and MAWPS-ComDiv datasets.
}
\label{tab:ablation_study}
\end{table}

To further investigate the mechanism of the proposed data augmentation method, we remove the data ranker, operation 1 and operation 2 in turn and test MathEn+DA on the Math23K-ComDiv and MAWPS-ComDiv datasets.
We show the results in Table~\ref{tab:ablation_study}.
As we can see, removing any of them causes the decrease of results.
This indicates that data ranker, operation 1 and operation 2 all contribute to the good effect of data augmentation method.
Operation 1 and operation 2 generate different compositions to fill the gap.
Data ranker prevents the model from the hurt of low-qualified augmented data.

\paragraph{Effect of Iteration Time.} 

\begin{figure*}
    \centering
    \includegraphics[scale=0.48, trim=0.95cm 0cm 0.75cm 0cm, clip]{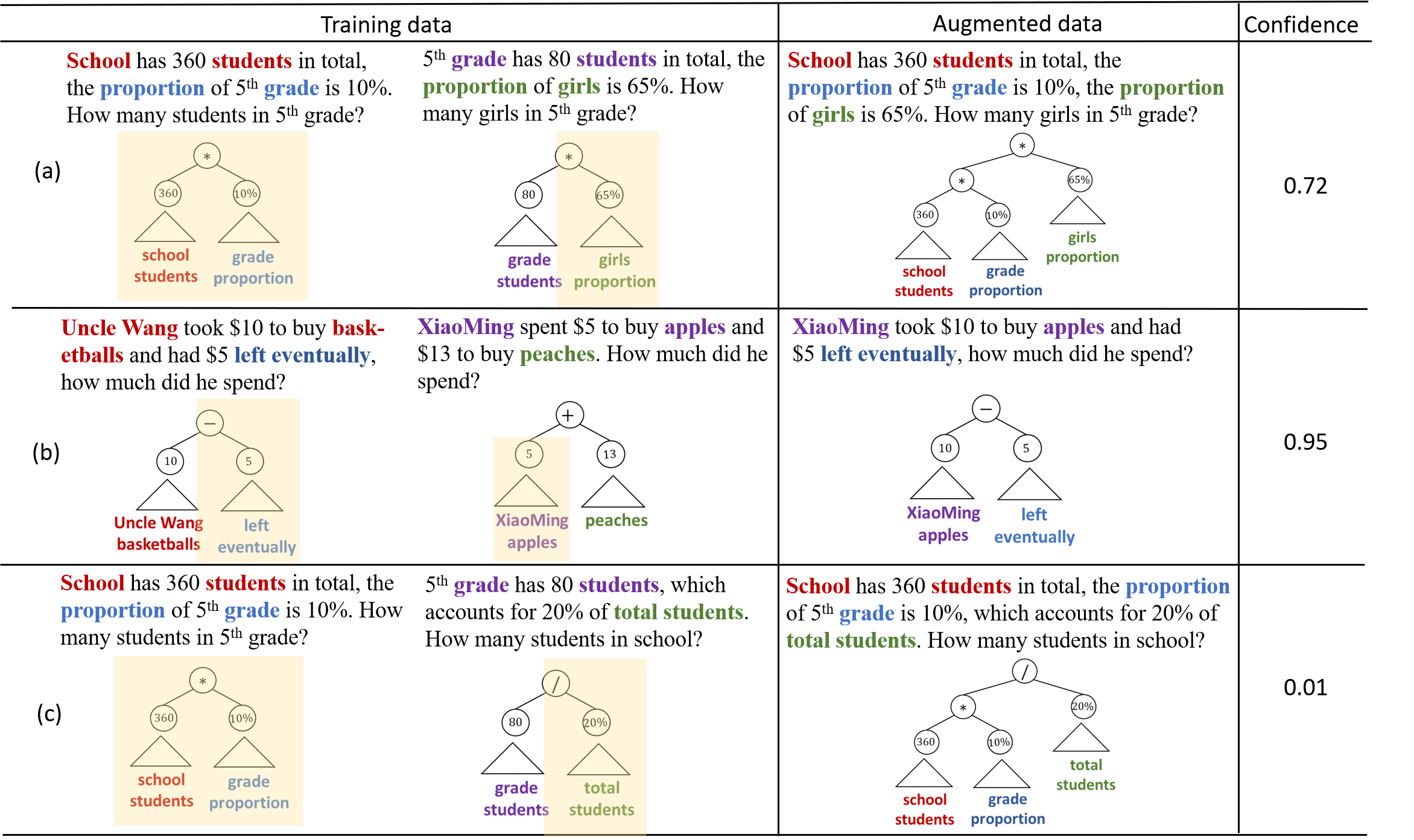}
    \caption{Three augmented examples produced via our method. The original training examples of Math23K is shown at the left column, corresponding augmented data is shown at the middle column and the confidence score for the augmented data is shown at the right column. Each MWP is associated with its expression tree. The context of a quantity in MWPs is highlighted with colors and the partial expression trees forming the augmented data is highlighted with yellow boxes.}
    \label{fig:case_study}
\end{figure*}


We depict the curve of data augmentation effect with the influence of iteration time in Figure~\ref{fig:compound_divergence} (b).
As we can see, with the increase of iteration time, the answer accuracy of GTS on SD dataset increases gradually.
It is because more iteration indicates more augmented data is generated to fill the compositional gap.
But when iteration comes to $6$, the increase becomes less significant.
Meanwhile, we notice that with the increase of iteration, the average time for each training epoch becomes larger.
It is because that the number of augmented data increases exponentially, which leads to a large time cost for training the model.
Therefore, it is a trade-off to select a suitable iteration time with the consideration of both performance gain and time cost.

\paragraph{Human Evaluation of Augmented Data}
In order to examine the quality of augmented data and understand how the augmented data improve general MWP methods, we invite two people to evaluate the quality of augmented MWPs. 
An augmented MWP will be labeled as qualified by human evaluators if it satisfies grammatical correctness, fluency, and logical consistency. 
Otherwise, it will be labeled as unqualified. 
For simplicity, we randomly sample 100 augmented MWPs and ask human evaluators to judge the quality of each MWP. 
We then calculate the agreement between evaluators using Cohen's k coefficient. 
Finally, the percentage of qualified augmented examples from two evaluators are 83\% and 84\% respectively, and their Cohen's k coefficient is 0.71, which indicates the data augmentation indeed generates a decent number of high-qualified data examples to strength the training procedure. 

\paragraph{Case Study of Augmented Data}
We display some augmented data in Figure~\ref{fig:case_study}. 
As we can see, the expression tree of augmented data is a novel combination of the expression trees of training data.
The math word problems are the corresponding narratives of the expression trees.
Example (a) shows that data augmentation method generates a novel combination of two familiar structures. 
The training data contains example of the structure $n_1 \times n_2$.
But the combination $n_1 \times n_2 \times n_3$ has never been observed in training.
The augmented data fills this gap.
After adding the augmented data into the training set, such combination rules will be learned by models.
Example (b) shows that data augmentation generates a novel combination of a familiar primitive and a familiar structure.
The training data contains example of the structure $n_1 - n_2$ but the generalization concerns the constituents "XiaoMing", "apples" has never been observed in the first position of $n_1 - n_2$.
The augmented data fills this gap by including more lexical variations into the training data and thus improve the lexical generalization of models.
Example (c) is another augmented example of structural generalization. 
Even though it is semantically correct, it is logically wrong.
When a data ranker generates its math expression, and the confidence score is very low.
Therefore, this is not a good augmented example and our method excludes it from the training set.



\section{Conclusion}

In this manuscript, we investigated compositional generalization in MWP solving task.
We created three compositional challenges on MWPs and evaluated a set of methods.
We observed most of the methods suffered from a big loss of performance.
We further proposed an iterative data augmentation method and it has proven to be effective in improving the compositional generalization of general methods.
While this problem is still open to be explored, our study provides some insights for proposing new solutions in the future research.

\section{Appendix}
\label{sec:appendix}

\subsection{Rule Table for Synthesizing Data}
We manually defined 13 rules in total to generate synthetic data (Sec.~\ref{sec:synthesize_method}) which are displayed in Table~\ref{tab:temp_comp}. 

\begin{table}[t]
    \centering
    \small
    \begin{tabular}{c |c |c | c c}
    \hline
    & ID & Seed MWP & \multicolumn{2}{c}{Synthetic MWP(s)} \\
    \hline \hline
    \parbox[t]{2mm}{\multirow{18}{*}{\rotatebox[origin=c]{90}{Decomposition}}} & \multirow{2}{*}{1} & $\{s_1, s_2, s_3\}$ & $\{s1, \Delta, s_3 \}$ & $\{s_2, \Delta\}$ \\
    & & $n_1 \times (1-n_2)$ & $n_1 \times n_2$ & $1-n_1$ \\
    \cline{2-5}
    & \multirow{2}{*}{2}& $\{s_1, s_2, s_3, s_4\}$ & $\{s_1, s_2, \Delta\}$ & $\{\Delta, s_3, s_4\}$ \\
    & & $n_1 \times n_2/n_3$ & $n_1 \times n_2$ & $n_1/n_2$ \\
    \cline{2-5}
    & \multirow{2}{*}{3} & $\{s_1, s_2, s_3, s_4\}$ & $\{s_1, s_2, \Delta\}$ & $\{\Delta, s_3, s_4 \}$\\
    & & $n_1 \times n_2 + n_3$ & $n_1 \times n_2$ & $n_1+n_2$ \\
    \cline{2-5}
    & \multirow{2}{*}{4} & $\{s_1, s_2, s_3, s_4\}$ & $\{s_1, s_2, \Delta \}$ & $\{\Delta, s_3, s_4\}$ \\
    & & $n_1 \times n_2 \times n_3$ & $n_1 \times n_2$ & $n_1 \times n_2$ \\
    \cline{2-5}
    & \multirow{2}{*}{5} & $\{s_1, s_2, s_3\}$ & $\{s_1, \Delta, s_3\}$ & $\{s_2, \Delta \}$ \\
    & & $n_1 / (1-n_2)$ & $n_1/n_2$ & $1-n_1$ \\
    \cline{2-5}
    & \multirow{2}{*}{6} & $\{s_1, s_2, s_3, s_4\}$ & $\{s_1, s_2, \Delta \}$ & $\{\Delta, s_3, s_4\}$ \\
    & & $(n_1 - n_2)/n_3$ & $n_1-n_2$ & $n_1/n_2$ \\
    \cline{2-5}
    & \multirow{2}{*}{7} & $\{s_1, s_2, s_3, s_4\}$ & $\{s_1, s_2, \Delta \}$ & $\{\Delta, s_3, s_4\}$ \\
    & & $(n_1+n_2) \times n_3$ & $n_1+n_2$ & $n_1 \times n_2$ \\
    \cline{2-5}
    & \multirow{2}{*}{8} & $\{s_1, s_2, s_3, s_4\}$ & $\{s_1, s_2, \Delta \}$ & $\{\Delta, s_3, s_4\}$ \\
    & & $n_1 /n_2 \times n_3$ & $n_1/n_2$ & $n_1 \times n_2$ \\
    \cline{2-5}
    & \multirow{2}{*}{9} & $\{s_1, s_2, s_3, s_4\}$ & $\{s_1, s_2, \Delta \}$ & $\{\Delta, s_3, s_4\}$ \\
    & & $n_1 \times n_2 - n_3$ & $n_1 \times n_2$ & $n_1-n_2$ \\
    \hline \hline
    \parbox[t]{2mm}{\multirow{8}{*}{\rotatebox[origin=c]{90}{Reformulation}}} & \multirow{2}{*}{10} & $\{s_1, s_2, s_3\}$ & $\{s_1, s_2, \Delta \}$ & \\
    & & $n_1 + n_2$ & $n_1 - n_2$ & \\
    \cline{2-5}
    & \multirow{2}{*}{11} & $\{s_1, s_2, s_3\}$ & $\{s_1, s_2, \Delta \}$ & \\
    & & $n_1 - n_2$ & $n_1 + n_2$ & \\
    \cline{2-5}    
    & \multirow{2}{*}{12} & $\{s_1, s_2, s_3\}$ & $\{s_1, s_2, \Delta \}$ & \\
    & & $n_1 \times n_2$ & $n_1 / n_2$ & \\
    \cline{2-5}    
    & \multirow{2}{*}{13} & $\{s_1, s_2, s_3\}$ & $\{s_1, s_2, \Delta \}$ & \\
    & & $n_1 / n_2$ & $n_1 \times n_2$ & \\
    \hline 
    \end{tabular}
    \caption{The rule table used in our work. 
    The middle section contains all the rules for decomposition and the third section contains all the rules for reformulation.}
    \label{tab:temp_comp}
\end{table}

\ignore{
\subsection{Atom and Compound Distributions}
\begin{figure}
\centering
    \begin{subfigure}[b]{.48\linewidth}          
        \includegraphics[scale=0.25]{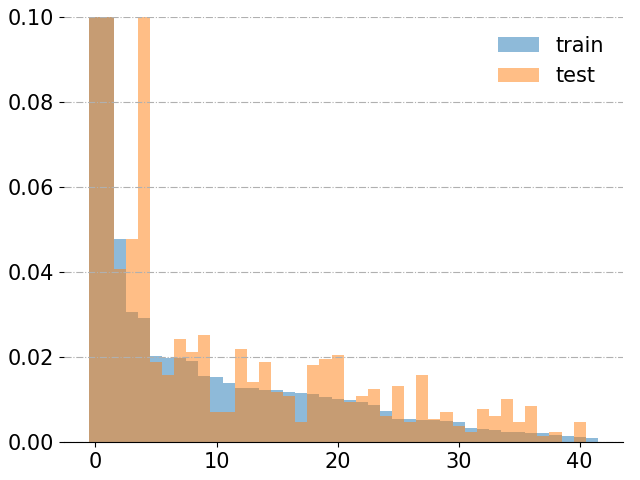}
        \caption{}
    \end{subfigure}%
    \hspace{0.1cm}
    \begin{subfigure}[b]{.48\linewidth}
        \includegraphics[scale=0.25]{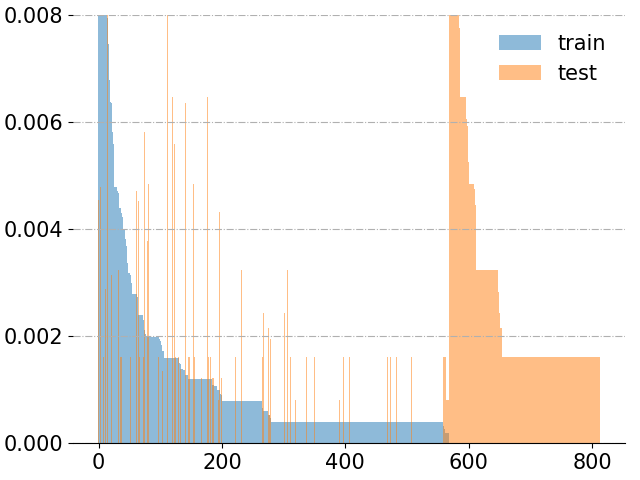}
        \caption{}
    \end{subfigure}
\caption{Distributions of (a) atoms and (b) compounds in the training vs. test set of SD data set.}
\label{fig:frequency}
\end{figure}

To verify that our compositional data sets could effectively retain the objective of maximizing $\text{DBCA}^c$ with a upper bound of $\text{DBCA}^a$, we display Figure~\ref{fig:frequency} to show the distribution of atoms and compounds in the training and test of SD data set when $k=40$ for better visualization.
As we can see, the atom frequency of these two sets are similar while the compounds are distributed differently.
}

\subsection{Hyperparameters Choice}

We show the hyperparameters of all the tested methods in Table~\ref{tab:hyperparameters}.  We set hyperparameters of any model to be the same on all datasets. For BERTGen and GPT-2, we use original default hyperparameters of BERT and GPT-2, respectively. We tune hyper-parameters, i.e., learning rate, batch size, dimension of latent vector (including embedding size and hidden size), dropout ratio, number of layers, and number of heads, within the range of \{0.0001, 0.0003, 0.0005, 0.001, 0.003, 0.01\}, \{16, 32, 64, 128\}, \{128, 256, 512\}, \{0.05, 0.1, 0.15, 0.2\}, and \{1,2,4,8\}, and \{2,4,8\} respectively and choose the best group of hyperparameters by validation accurarcy.

\begin{table}[t]
    \centering \small
    \scalebox{0.96}{
    \begin{tabular}{@{}lcccc@{}}
    \hline
         &  LSTM & Transformer & MathEN & GTS\\
        \hline \hline 
        train epochs  & 100 &  100 & 100 & 100\\
        batch size  &32 & 64 & 32 & 64\\
        embedding size &128 & 512 & 128 & 128 \\
        hidden size  & 512 & 512 & 512 & 512\\
        \# layers  & 2 & 4 & 2 &2\\
        \# heads   & -- & 8 & -- & --\\
        learning rate  & 0.001 & 0.003  & 0.001 & 0.001\\
        warmup steps & -- & 1500 & -- & --\\
        dropout & 0.5 & -- & 0.5 & 0.5\\
        \hline
    \end{tabular}
    }
    \caption{Summary of hyperparameters that deviate from the defaults.}
    \label{tab:hyperparameters}
\end{table}

\bibliography{reference}
\bibliographystyle{IEEEtran}

\end{document}